\def\year{2022}\relax
\documentclass[letterpaper]{article} 
\usepackage{aaai22}  
\usepackage{times}  
\usepackage{helvet}  
\usepackage{courier}  
\usepackage[hyphens]{url}  
\usepackage{graphicx} 
\usepackage{natbib}  
\usepackage{caption} 

\usepackage{amsmath}
\usepackage{amssymb}
\usepackage{multirow,makecell,bbm}

\DeclareCaptionStyle{ruled}{labelfont=normalfont,labelsep=colon,strut=off} 
\usepackage{algorithm}
\usepackage{algorithmic}

\usepackage{newfloat}
\usepackage{listings}
\floatstyle{ruled}
\newfloat{listing}{tb}{lst}{}
\floatname{listing}{Listing}

\title{Learning to Predict 3D Lane Shape and Camera Pose  from a Single Image via Geometry Constraints}
\author {
    Ruijin Liu\textsuperscript{\rm 1},
    Dapeng Chen\textsuperscript{\rm 2}, 
    Tie Liu\textsuperscript{\rm 3}, 
    Zhiliang Xiong\textsuperscript{\rm 4},
    Zejian Yuan\textsuperscript{\rm 1}
}
\affiliations {
    \textsuperscript{\rm 1} Institute of Artificial Intelligence and Robotics, Xi’an Jiaotong University, Xi'an, China\\
    \textsuperscript{\rm 2} The Hong Kong Polytechnic University, Hong Kong SAR, China\\
    \textsuperscript{\rm 3} College of Information Engineering, Capital Normal
    University, Beijing, China\\
    \textsuperscript{\rm 4} Shenzhen Forward Innovation Digital Technology Co. Ltd, Shenzhen, China\\
    lrj466097290@stu.xjtu.edu.cn, dapengchenxjtu@yahoo.com, yuan.ze.jian@xjtu.edu.cn, liutiel@163.com, leslie.xiong@forward-innovation.com
}

\usepackage{bibentry}

\begin{document}

\maketitle

\begin{abstract}
Detecting 3D lanes from the camera is a rising problem for autonomous vehicles. In this task,  the correct camera pose is the key to generating accurate lanes, which can transform an image from perspective-view to the top-view. With this transformation, we can get rid of the perspective effects so that 3D lanes would look similar and can accurately be fitted by low-order polynomials. However, mainstream 3D lane detectors rely on perfect camera poses provided by other sensors, which is expensive and encounters multi-sensor calibration issues.  To overcome this problem, we propose to predict 3D lanes by estimating camera pose from a single image with a two-stage framework. The first stage aims at the camera pose task from perspective-view images. To improve pose estimation, we introduce an auxiliary 3D lane task and geometry constraints to benefit from multi-task learning, which enhances consistencies between 3D and 2D, as well as compatibility in the above two tasks. The second stage targets the 3D lane task. It uses previously estimated pose to generate top-view images containing distance-invariant lane appearances for predicting accurate 3D lanes. Experiments demonstrate that, without ground truth camera pose, our method outperforms the state-of-the-art perfect-camera-pose-based methods and has the fewest parameters and computations. Codes are available at \url{https://github.com/liuruijin17/CLGo}.
\end{abstract}

\section{Introduction}

\noindent Compared with 2D lane detection, image-based 3D lane detection is beneficial to perceiving a real-world driving environment, which is crucial for intelligent cruise control, high definition map construction, and traffic accident reduction in autonomous driving~\cite{HDMaps, Uncertainty}. Unlike the  2D lane detection that relies on the flat ground plane assumption, the 3D method is more flexible to handle complex road undulations. It usually requires a camera pose to transform the image/features of the perspective view to the top-view by inverse perspective mapping. The current methods utilize the ground truth camera pose provided by the benchmark to estimate the accurate top-view feature. Such a strategy is potentially expensive in realistic driving applications because it needs an additional third-party tool (inertial sensor or SfM algorithm~\cite{SfM,IMU}) to provide an accurate camera pose steadily during driving~\cite{Gen3rdTool,GenLaneNet}

Instead of utilizing the ground truth camera pose, we propose to learn camera pose online to perform view transform for 3D lane detection and impose geometry consistency constraints to improve the pose results. Specifically, we utilize a two-stage framework to learn from different viewpoints. The first stage aims at learning camera pose from perspective-view images. To improve the camera pose, we introduce geometry constraints. The constraints are based on an auxiliary network branch that predicts the 3D lanes. With the predicted camera pose, the predicted 3D lanes are projected to the camera plane and the virtual flat ground plane, which can be supervised by the ground truth lanes of the two planes. The second stage uses transformed top-view images based on learned camera pose to detect ultimate 3D lanes. In the top-view, lanes have a similar appearance along a longitudinal direction which is crucial for deciphering 3D lanes, especially in distant areas. The 3D lanes are modeled by two polynomials: one approximates the lateral variations, and the other expresses undulation changes along longitudinal positions. To better capture lanes' long and thin structures, both stages utilize transformers~\cite{AttentionIsAllYourNeed} to aggregate non-local context.

Our method is evaluated on the only public synthetic 3D lane detection benchmark and a self-collected real-world dataset which will be released. Without utilizing the ground truth camera poses, our method outperforms previous state-of-the-art methods that need ground truth camera poses. Furthermore, our approach has a light model size, few computation costs, and quite fast speed, showing great potential in real driving applications. The main contributions are summarized as follows:  
\begin{itemize}
 \item We design a two-stage framework that firstly predicts the camera pose, then utilizes the camera pose to generate a top-view image for accurate 3D lane detection. 
\item We propose geometry constraints to assist in estimating camera pose, which enforces the consistency between the predicted 3D poses and lanes with the ground truth lanes on the 2D plane. 
\item We employ polynomials to model the 3D lane, which preserves the lanes' local smoothness and global shape and avoids the complex post-processing. 
\end{itemize}

\section{Related Work}

The field of vision-based lane detection has grown considerably in the last decade. The popularity of camera sensors has allowed lane detection in 2D to gain significant momentum. Traditional methods typically design hand-crafted features, adopt mathematical optimized algorithms and geometry or scene context to learn lane structures well~\cite{LDWReview,LaneFeatureMethod,LaneModelMethod}. Deep methods built by convolutions have exploded in recent years, making significant progress and applicable systems for real applications~\cite{dpcheng3,zhaoyun1}. Two-stage methods which extract segmentation or proposal plus post-processing ruled the filed for several years~\cite{LaneNet,FastDraw,SCNN,LaneAndRoad,ENet-SAD,PINet,Line-CNN,LaneATT,UltraFast,CurveLaneNAS,LightWeightECCV,RowwiseClassification,HESA,RESA}. To streamline the pipeline into an end-to-end fashion, single-stage methods~\cite{PolyLaneNet,LSTR} directly estimate coefficients of prior mathematical curves have shown both higher efficacy and efficiency. However, those 2D detectors are built with specific planar world assumptions, resulting in a limited representation of realistic and complex 3D road undulations.

The 3D lane detection task has attracted research interest because it doesn't rely on the plane assumption to predict the lanes~\cite{GenRef5,GenRef19,GenRef2,3DLaneNet,GenLaneNet,Uncertainty,3DLaneICIP}. However, detecting a 3D lane from a single RGB image is non-trivial. The results can be easily affected by appearance variation due to changes in lighting condition, texture, background, and camera pose.  One solution is to utilize the extra information from other sensors such as LiDAR or stereo cameras~\cite{GenRef5,GenRef19,GenRef2}. Still, the cost of these devices is too expensive to be applied to them on consumer-grade products widely~\cite{GenLaneNet}, while arising multi-sensor calibration problems during driving. 
Unlike multi-sensor methods, monocular methods~\cite{3DLaneNet,GenLaneNet, Uncertainty} only require a single image and camera pose to transform image/features from perspective-view to top-view for accurate 3D lane detection. Previous methods rely on the ground truth camera pose for evaluation and cannot work well when the camera pose changes dynamically when driving at rough terrain or accelerating~\cite{MonoEF}). In contrast, our method provides a full-vision-based 3D lane detector that is not dependent on the ground truth camera pose and other sensors.  It can flexibly adapt to the changing driving environment with an affordable device cost.

\section{Methodology}

\subsection{3D Lane Geometry}
\label{subsec:geometry}

\begin{figure}[t]
\begin{center}
\includegraphics[width=83mm]{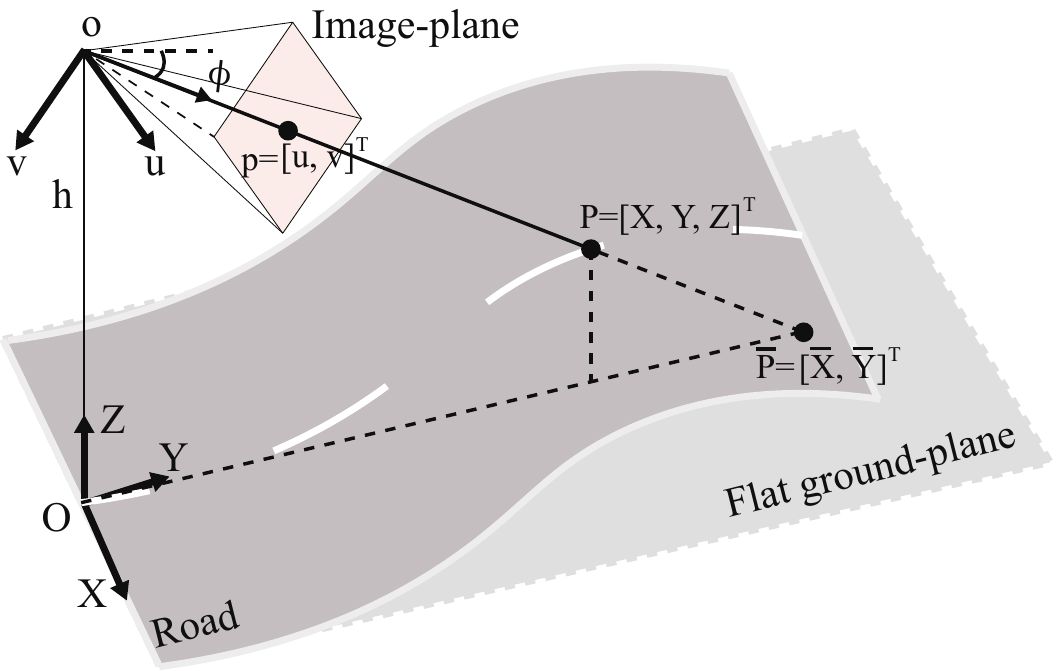}
\end{center}
\caption{\textbf{Geometry setup} about camera and 3D lane.}
\label{fig:geometry}
\end{figure}

We first define notations for the lane representation in the 3D space, perspective image plane, and a virtual flat ground plane as shown in Fig. \ref{fig:geometry} . Then we introduce the transformations between these spaces. 

\noindent \textbf{Notation of Lane.}
Suppose $\mathbf{P}=\left[X, Y, Z\right]^T$ is a lane point in the 3D space.  Its projection on the flat ground-plane (i.e.,  the plane with Z=0 in the 3D space) is denoted by $\bar{\mathbf{P}}=\left[\bar{X}, \bar{Y}\right]^T$, while its projection on the camera plane is denoted by $\mathbf{p}=\left[u, v\right]^T$.  In the 3D space,  the origin of the coordinate $\mathbf{O}$ is the perpendicular projection of the camera center $\mathbf{o}$ on the plane $Z=0$. For the camera, we fix the intrinsic parameters $f_x, f_y, c_x$ and $ c_y$, and only estimate the camera height $h$ and pitch angle $\phi$ following the common setup~\cite{3DLaneNet,GenLaneNet}.

\noindent \textbf{Geometric transformation.}
The geometric transformation projects points of the lanes from the 3D space to the flat ground plane. In particular, a point $\mathbf{P}$ in the 3D space should be on the same line with the camera center $\mathbf{o}$ and the projected point $\bar{\mathbf{P}}$ on the flat ground plane. Therefore, we have $\frac{h}{h-Z}=\frac{\bar{X}}{X}=\frac{\bar{Y}}{Y}$. Note that it holds no matter $Z$ is positive or negative. The geometric transformation from 3D space to the flat ground-plane is written as:
\begin{equation}
\label{eq:gflat2g}
\begin{bmatrix}
   \bar{X} \\
   \bar{Y} \\
\end{bmatrix}
=
\begin{bmatrix}
   \frac{h}{h-Z} &0 &0 \\
   0 &\frac{h}{h-Z} &0 \\
\end{bmatrix}
\begin{bmatrix}
   X \\
   Y \\
   Z
\end{bmatrix},
\end{equation}
For the simplicity of the notation, Eq. \ref{eq:gflat2g} is also formulated as
$\bar{\mathbf{P}}=\mathbf{G}\mathbf{P}$, and $\mathbf{G}$ is the geometric transformation matrix. 

\noindent \textbf{Homography transformation.} 
The homography transformation projects the point from the flat ground plane to the image plane.  Given a point $\bar{\mathbf{P}}$ on the flat ground plane and its corresponding point $\mathbf{p}$ on the image plane, the homography transformation are performed with their homogeneous coordinates:
\begin{equation}
\label{eq:gflat2img}
\begin{bmatrix}
   \tilde{u} \\
   \tilde{v} \\
   \tilde{z}
\end{bmatrix}
=
\begin{bmatrix}
   f_x &0 &c_x \\
   0 &f_y &c_y \\
   0 &0 &1
\end{bmatrix}
\begin{bmatrix}
   1 &0 &0 \\
   0 &\cos \left(\phi+\frac{\pi}{2}\right) &h \\
   0 &\sin \left(\phi+\frac{\pi}{2}\right) &0
\end{bmatrix}
\begin{bmatrix}
   \bar{X} \\
   \bar{Y} \\
   1
\end{bmatrix},
\end{equation}
We denote $\tilde{\mathbf{p}}=\left[\tilde{u},\tilde{v},\tilde{z}\right]^T$ and $\tilde{\mathbf{P}}=\left[\bar{X},\bar{Y},1\right]^T$ as the homogeneous  coordinates of $\mathbf{p}$ and $\bar{\mathbf{P}}$, therefore $u=\tilde{u}/\tilde{z}, v=\tilde{v}/\tilde{z}$. For simplicity, Eq. \ref{eq:gflat2img} is also formulated as $\tilde{\mathbf{p}} = \mathbf{H} \tilde{\mathbf{P}}$, and $\mathbf{H}$ is the Homography transformation matrix.

\begin{figure*}[t]
\begin{center}
\includegraphics[width=170mm]{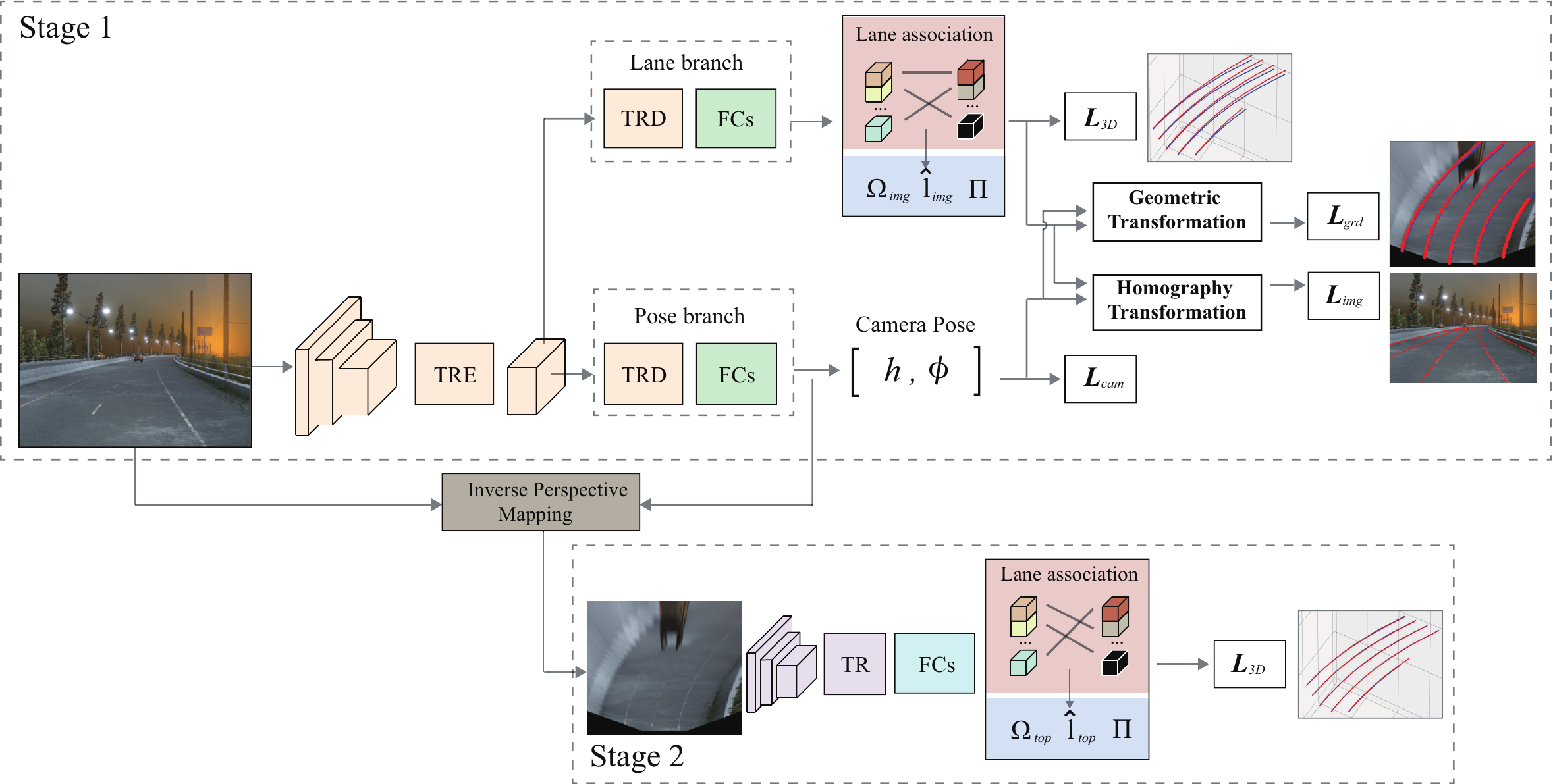}
\end{center}
\caption{\textbf{System Overview.} Stage 1 learns camera pose with the help of an auxiliary lane branch and geometry constraints. Then, the estimated camera pose transforms image from perspective-view into top-view where lanes look similar. Finally, Stage 2 aims at predicting ultimate 3D lanes from distance-invariant top-view image accurately.}
\label{fig:twostages}
\end{figure*}

\subsection{Output Definition}
Our method outputs both the camera pose and the detected lane in the 3D space. 

\noindent \textbf{Camera pose}.
As stated in the aforementioned geometry setting, the output camera pose contains the camera height $h$ and camera pitch angle $\phi$.

\noindent \textbf{3D Lane}. We use two polynomials to represent a 3D lane. For any point $[X,Y,Z]$ on the lane, $X,Z$ can be represented as polynomial functions of $Y$, that is:
\begin{equation}
\label{eq:3dshape}
\begin{bmatrix}
   X \\
   Z
\end{bmatrix}
=
\begin{bmatrix}
   \sum_{r=0}^R a_rY^r \\
   \sum_{r=0}^R b_rY^r \\
\end{bmatrix},
\end{equation}
where $a_R \neq 0$, $b_R \neq 0$ and $a_r, b_r$ are real number coefficients. $R$ indicates the order of the polynomial. Let $t_{1n}$ and $t_{2n}$ be the upper bound and lower bound of $Y$ of the $n$-th lane, the $n$-th lane in an image can be represented by the coefficients of the polynomial functions and the bounds of $Y$:
\begin{equation}
\label{eq:3dmodel}
\theta_n=\left\{\left\{a_{rn}, b_{rn}\right\}_{r=0}^R, t_{1n}, t_{2n}\right\},
\end{equation}
where $n \in \left\{1,...,N\right\}$ and $N$ is the number of ground truth 3D lanes in the image. 

\subsection{Network Architecture}

Fig.~\ref{fig:twostages} illustrates the overall two-stage structures. Stage 1 contains a backbone for feature extraction over the entire image and a transformer-encoder (TRE) that aggregates non-local relations among spatial features. The output of TRE is fed to two branches: a pose branch that can decode ego camera pose and a lane branch for 3D lanes.

With an estimated camera pose, \textit{inverse perspective mapping} (IPM) transforms the image into a top-view image. Then stage 2 employs a similar transformer-based network to be specialized at only predicting 3D lanes since lifting to top-view space makes the 3D lane fitting a much more simple task due to distance-invariant appearance features.

\noindent \textbf{Stage 1.}
Given an image, stage 1 extracts convolution features, then aggregates spatial features with a backbone and transformer-encoder just like LSTR~\cite{LSTR}. After obtaining the encoded feature, the pose branch decodes the camera pose (camera height $h$ and pitch angle $\phi$ ) of the current image. To help the camera pose learning, we introduce a lane branch to decode 3D lanes. The 3D lanes are then projected to the flat ground plane and the image plane by the predicted camera pose, which are supervised by the ground truth lanes on the two planes.

\noindent \textbf{Stage 2.} 
Stage 2 mainly follows the pipeline in stage 1 with the only lane branch to estimate 3D lanes, which is the ultimate detection results for evaluation. Different from stage 1,  stage 2 only adopts the transformed top-view images where lanes have a similar shape, scale and appearance to help reconstruct 3D lanes accurately.

\subsection{Loss functions with Geometry Constraints}

We train the network of Stage 1 with geometry constraints by using $L_{s1} = L_{cam}+L_{3D}+L_{grd}+L_{img}$, where $L_{3D}$ and $L_{cam}$ are basic fitting losses to supervise the 3D lane and the camera pose, respectively. While $L_{img}$ and $L_{grd}$ serve as geometry constraints that assist the learning of camera pose. In stage 2, only $L_{3D}$ is employed to train the network for more accurate 3D lane detection. We now introduce these loss functions one by one.

\noindent \textbf{Camera pose regression loss.}
The regression loss for camera pose has the form of: $L_{cam} = \left|\hat{h}-h\right| + \left|\hat{\phi}-\phi\right|$, where $\left|\cdot\right|$ is the mean absolute error.

\noindent \textbf{3D lane fitting loss.}
The loss is used to supervise the lane in the 3D space. It is noteworthy that the lane branch outputs the polynomial coefficients of $M$ 3D lanes,  where $M$ is larger than the maximum number of labeling lanes in the dataset. Because the loss function does not know the association between the predicted lanes and the ground truth lanes, we first associate the predicted curves and ground truth lanes by solving a bipartite matching problem.

\emph{3D lane association.} Let  $\Omega=\left\{\omega_m|\omega_m=\left(c_m, \theta_m\right)\right\}_{m=1}^{M}$ be
the set of the predicted 3D lanes, where $c \in \left\{0, 1\right\}$ (0: none-lane, 1: lane), and  $\theta_m=\left\{\left\{a_{rm}, b_{rm}\right\}_{r=0}^R, t_{1m}, t_{2m}\right\}$. Let $\Pi=\left\{\hat{\pi}_m|\hat{\pi}_m=\left(\hat{c}_m,\hat{\mathbf{L}}_m\right)\right\}_{m=1}^M$ be the set of ground truth 3D lanes, where the $m$-th ground truth lane $\hat{\mathbf{L}}_m=\left[\hat{X}_{km},\hat{Y}_{km},\hat{Z}_{km}\right]_{k=1}^{K}$. Note that $\Pi$ is padded with non-lanes to fill enough the number of ground truth lanes to $M$ for associating with $\Omega$. Next, we form a bipartite matching problem between $\Omega$ and $\Pi$ to build lane association.
The problem is formulated as a distance minimizing task by searching an injective function $l: \Pi \rightarrow \Omega$, where $l\left(m\right)$ is the index of a 3D lane prediction $\omega_{l\left(m\right)}$ which is assigned to $m$-th ground truth 3D lane $\hat{\pi}_m$:

\begin{equation}
\label{optimalproblem}
\hat{l}=\mathop{\arg\min}_{l} \sum_{m=1}^M D\left(\hat{\pi}_m, \omega_{l\left(m\right)}\right),
\end{equation}
based on the matching cost:
\begin{equation}\small
\label{matchingcost}
\begin{split}
& D =  - \alpha_1 p_{l\left(m\right)}\left(\hat{c}_m\right) + \mathbbm{1}\left(\hat{c}_m = 1\right) \alpha_2  \left|\hat{\mathbf{L}}_m - \mathbf{L}_{l\left(m\right)} \right| \\
& + \mathbbm{1}\left(\hat{c}_m = 1\right) \alpha_3 \left(\left|\hat{Y}_{1m} - t_{1l\left(m\right)}\right| + \left|\hat{Y}_{Km} - t_{2l\left(m\right)} \right|\right),
\end{split}
\end{equation}
where $\alpha_1$, $\alpha_2$ and $\alpha_3$ are coefficients which adjust the loss effects of classification, polynomials fitting and boundaries regression, and $\mathbbm{1}\left(\cdot\right)$ is an indicator function. The $l\left(m\right)$-th prediction lane $\mathbf{L}_{l\left(m\right)}=\left[X_{kl\left(m\right)}, Y_{kl\left(m\right)}, Z_{kl\left(m\right)}\right]_{k=1}^K=\left[\sum_{r=0}^R a_{rl\left(m\right)} \hat{Y}_{km}^r,~~\hat{Y}_{km},~~\sum_{r=0}^R b_{rl\left(m\right)} \hat{Y}_{km}^r\right]_{k=1}^K$.

\emph{3D lane fitting.} After getting the optimized injective function $\hat{l}$ by solving Eq.~\ref{optimalproblem} using Hungarian algorithms~\cite{DETR}, the 3D fitting loss can be defined as:
\begin{equation}\small
\label{lossin3d}
\begin{split}
& L_{3D} = \sum_{m=1}^M \mathbbm{1}\left(\hat{c}_m = 1\right) \alpha_3 \left(\left|\hat{Y}_{1m} - t_{1\hat{l}\left(m\right)}\right| + \left|\hat{Y}_{Km} - t_{2\hat{l}\left(m\right)}\right|\right)
\\
& +   \mathbbm{1}\left(\hat{c}_m = 1\right) \alpha_2 \left|\hat{\mathbf{L}}_m - \mathbf{L}_{\hat{l}\left(m\right)} \right|  -\alpha_1 \log p_{\hat{l}\left(m\right)}\left(\hat{c}_m\right).
\end{split}
\end{equation}
where $\alpha_1$, $\alpha_2$ and $\alpha_3$ are the same coefficients with Eq.~\ref{matchingcost}.

\begin{table*}[t]
\begin{center}
\setlength{\tabcolsep}{2.0mm}{
\begin{tabular}{|c|l|c|c|c|c|c|c|c|c|c|}
\hline
Scene & Method & CP & Height & Pitch & F-Score & AP & X error near & X error far & Z error near & Z error far \\
\hline
\hline
\multirow{4}*{\makecell[c]{BS}} & 3D-LaneNet & GT & - & -
& 86.4 & 89.3 & 0.068 & 0.477 & 0.015 & \textbf{0.202} \\
~ & Gen-LaneNet & GT & - & - 
& 88.1 & 90.1 & \textbf{0.061} & 0.496 & \textbf{0.012} & 0.214 \\
~ & Stage 1~(ours) & $\backslash$  & 0.021 & $\mbox{0.121}^\circ$ 
& 88.2 & 90.3 & 0.092 & 0.507 & 0.038 & 0.277 \\
~ & CLGo~(ours) & PD & 0.021 & $\mbox{0.121}^\circ$ 
& \textbf{91.9} & \textbf{94.2} & \textbf{0.061} & \textbf{0.361} & 0.029 & 0.250 \\
\hline
\hline
\multirow{4}*{\makecell[c]{ROS}} & 3D-LaneNet & GT & - & - 
& 72.0 & 74.6 & 0.166 & 0.855 & 0.039 & \textbf{0.521} \\
~ & Gen-LaneNet & GT & - & - 
& 78.0 & 79.0 & \textbf{0.139} & 0.903 & \textbf{0.030} & 0.539 \\
~ & Stage 1~(ours) & $\backslash$ & 0.042 & $\mbox{0.303}^\circ$ 
& 77.1 & 78.9 & 0.193 & 0.919 & 0.077 & 0.679 \\
~ & CLGo~(ours) & PD & 0.042 & $\mbox{0.303}^\circ$ 
& \textbf{86.1} & \textbf{88.3} & 0.147 & \textbf{0.735} & 0.071 & 0.609 \\
\hline
\hline
\multirow{4}*{\makecell[c]{SVV}} & 3D-LaneNet & GT & - & - 
& 72.5 & 74.9 & 0.115 & 0.601 & 0.032 & \textbf{0.230}    \\
~ & Gen-LaneNet & GT & - & - 
& 85.3 & 87.2 & \textbf{0.074} & 0.538 & \textbf{0.015} & 0.232 \\
~ & Stage 1~(ours)  & $\backslash$ & 0.026 & $\mbox{0.155}^\circ$ 
& 84.2 & 85.9 & 0.117 & 0.519 & 0.044 & 0.317 \\
~ & CLGo~(ours) & PD & 0.026 & $\mbox{0.155}^\circ$ 
& \textbf{87.3} & \textbf{89.2} & 0.084 & \textbf{0.464} & 0.045 & 0.312 \\
\hline
\end{tabular}}
\caption{Comparisons on 3D lane synthetic dataset testing set~(\%). The Height error goes in \textbf{centimeters} and X error and Z error are given in \textbf{meters}. Pitch error is given in angular $\circ$. CP, GT and PD are abbreviations of camera pose, ground truth and prediction. The best results are in bold.}
\label{tab:OverallResult}
\end{center}
\end{table*}

\noindent \textbf{Geometry constraint in flat ground plane.}
Projecting lanes to the flat ground plane provides global shape patterns correlated with varying lane heights. If lanes in 3D space have no undulation, their projections on this plan will show parallel alignment and the same curvature variation. However, when encountering an uphill road, lane projections will not be parallel but converging in the bottom (or diverging for a downhill road). Such shape patterns are clearer to guide the network to learn 3D lanes and camera heights in fine detail. With the same optimal $\hat{l}$,  the fitting loss in the flat ground-plane can be written as:
\begin{equation}
\label{lossingflat}
\begin{split}
L_{grd} = \sum_{m=1}^M \mathbbm{1}\left(\hat{c}_m = 1\right) \alpha_2 \left|\hat{\bar{\mathbf{L}}}_m - \bar{\mathbf{L}}_{\hat{l}\left(m\right)} \right|.
\end{split}
\end{equation}
Here, $\hat{\bar{\mathbf{L}}}_m=\left[\hat{\bar{X}}_{km},\hat{\bar{Y}}_{km}\right]_{k=1}^K$ is the $m$-th ground truth projected lane.  It is obtained by projecting the 3D ground truth lane by geometric transformation (Eq.~\ref{eq:gflat2g}), where the geometric matrix is calculated by the ground truth camera height $\hat{h}$. Similarly, we can obtain the $\hat{l}\left(m\right)$-th predicted projected lane $\bar{\mathbf{L}}_{\hat{l}\left(m\right)}=\left[\bar{X}_{k\hat{l}\left(m\right)},\bar{Y}_{k\hat{l}\left(m\right)}\right]_{k=1}^K$ from the predicted 3D lane. Different from the ground truth projected lane, the geometric transformation $\mathbf{G}$ is built by the estimated camera height $h$ to transform 3D lanes onto the flat ground.

\noindent \textbf{Geometry constraint in image plane.}
Projecting lanes to this plane not only provides a supervision signal that is aligned with the input 2D image for free but also makes the 3D lanes and camera pose geometrically consistent with each other. More importantly, a small 3D jitter could cause a significant 2D shift. Such an amplification phenomenon undoubtedly facilitates the further optimization of 3D outputs. With the same optimal $\hat{l}$, the fitting loss in the image-plane is:
\begin{equation}
\label{lossinimg}
\begin{split}
L_{img} = \sum_{m=1}^M \mathbbm{1}\left(\hat{c}_m = 1\right) \alpha_2 \left|\hat{\mathbf{l}}_m- \mathbf{l}_{\hat{l}\left(m\right)} \right|.
\end{split}
\end{equation}
In particular, $\hat{\mathbf{l}}_m=\left[\hat{u}_{km},\hat{v}_{km}\right]_{k=1}^K$ is the $m$-th ground truth lane in the image plane. It is obtained by projecting the ground truth lane on the flat ground plane to the image plane via the homography transformation, where the transformation matrix constructed by the ground truth camera pose according to   
Eq.~\ref{eq:gflat2img}. Similarly, the $\hat{l}\left(m\right)$-th prediction $\mathbf{l}_{\hat{l}\left(m\right)}=\left[u_{k\hat{l}\left(m\right)},v_{k\hat{l}\left(m\right)}\right]_{k=1}^K$ is calculated by the predicted 3D lane and homography transformation matrix $\mathbf{H}$ based on the estimated camera pose $h$ and pitch angle $\phi$.

\noindent \textbf{Multi-task joint training strategy.} During training, we apply a multi-step training strategy to make optimization more stable. Specifically, we first train stage 1 and stage 2 separately, while stage 2 is fed with ground truth camera pose. Next, the whole architecture is jointly trained sequentially by feeding camera pose from stage 1 to stage 2. The IPM is a differentiable operation that can perform the transformation for either images or features. Thus, we could share parameters for two stages by applying IPM on features, enabling the part of the network before IPM to be reused for two stages to seek a better accuracy-cost trade-off. \textit{Detailed studies can be found at the appendix.}

\begin{figure*}[t]
\begin{center}
\includegraphics[width=160mm]{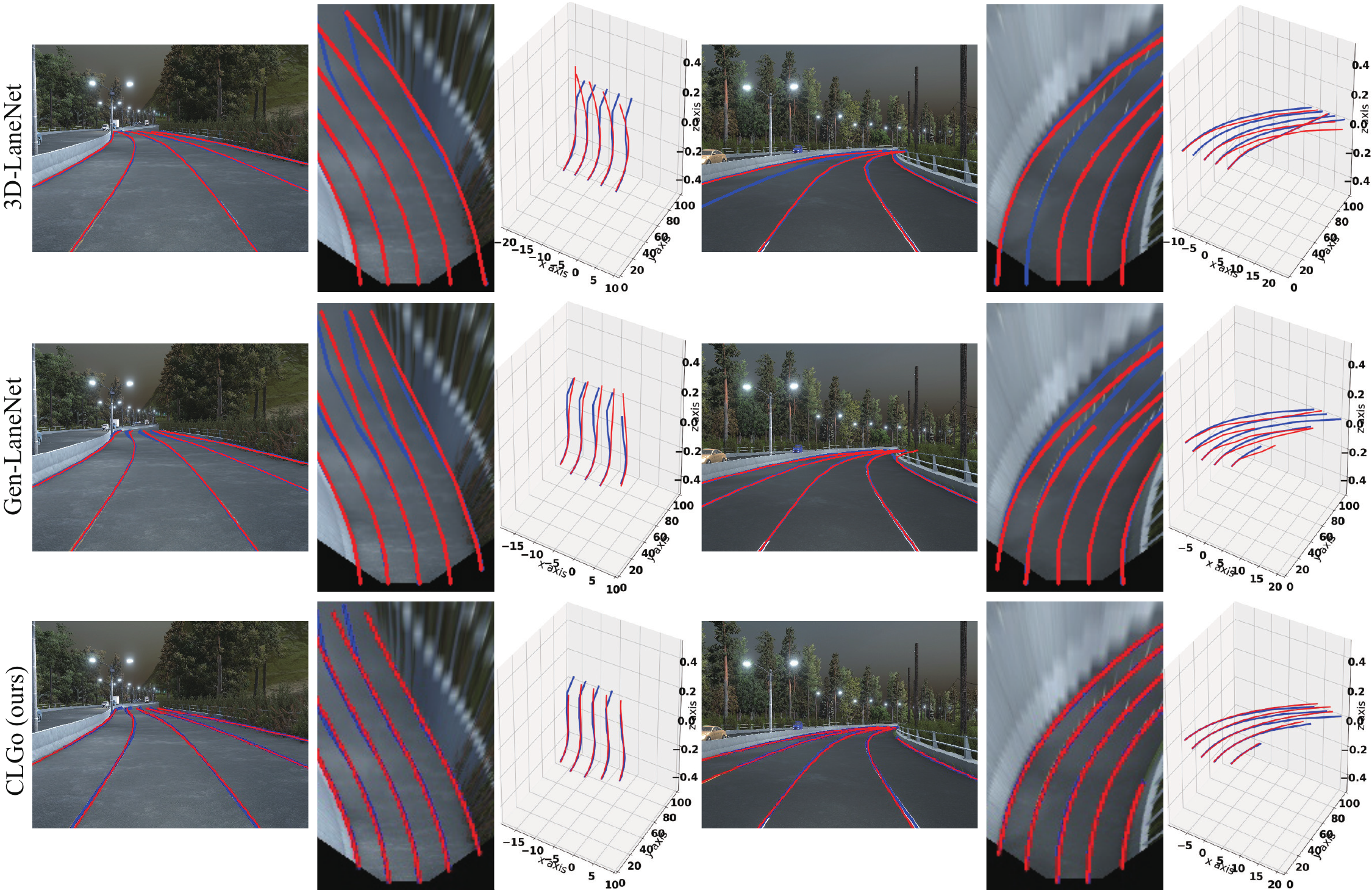}
\end{center}
\caption{\textbf{Qualitative comparisons} with 3D-LaneNet~\cite{3DLaneNet} and Gen-LaneNet~\cite{GenLaneNet} on the test set of balanced scenes. Red and blue lines indicate the estimation and ground truth lanes respectively. From right to left, we sequentially show the 3D fitting results and their projections on the flat ground plane and image plane.}
\label{fig:qual3dlane}
\end{figure*}

\begin{table}[t]
\begin{center}
\setlength{\tabcolsep}{3.6mm}{
\begin{tabular}{|l|c|c|c|c|}
\hline
Method       & FPS & MACs  & \#Para & PP \\
\hline\hline
3D-LaneNet   & 53  & 60.47 & 20.6 & \checkmark \\
Gen-LaneNet  & 60  & 9.85  & 3.4  & \checkmark \\
CLGo~(ours)         & \textbf{75} & \textbf{0.497}  & \textbf{1.5} & $\times$          \\
\hline
\end{tabular}}
\caption{Comparisons of resource consumption. The number of MACs and parameters (\#Para) are given in GHz and million. The PP means the requirement of post processing like outlier removal and non-maximum suppression.}
\label{tab:ResourceResult}
\end{center}
\vspace{-0.5em}
\end{table}

\section{Experiments}

\noindent \textbf{Datasets.}
We adopt the \textbf{ONLY} public 3D lane detection benchmark named 3D Lane Synthetic Dataset~\cite{GenLaneNet}. The dataset consists of about 10,500 high-quality $1080\times1920$ images, containing abundant visual elements built by the unity game engine.  The dataset exhibits highly diverse 3D worlds with realistic scenes across highways, urban and residential areas within the silicon valley in the United States under different weather conditions (morning, noon, evening). The camera intrinsic parameters are fixed, and the camera extrinsic parameters only differ in camera heights and camera pitch angles, which have the range of 1.4-1.8 meters and $0^\circ-10^\circ$. The dataset is split originally into three different scenes: (1) Balanced Scenes (BS), (2) Rarely Observed Scenes (ROS), and (3) Scenes with Visual Variations (SVV). The SVV scene is used to conduct ablation studies since it covers illumination changes that affect camera pose estimation significantly.

Since there is no public realistic dataset for 3D lane evaluation, we provide a self-collected dataset named Forward-view Lane and Camera Pose (FLCP) to be the first one. It contains 1,287 images, each of which has corresponding 2D lane labels and calibrated camera poses. The test pipeline is to apply 3D detectors on FLCP images and project 3D detections into 2D results based on camera poses for evaluations.

\noindent \textbf{Evaluation Metric.}
For 3D Lane Synthetic Dataset, we use the standard evaluation metric designed by Gen-LaneNet~\cite{GenLaneNet}. Given the prediction set and ground truth set, an edit distance $d_{edit}$ is defined to measure the lane-to-lane matching cost. 
For each prediction lane, it will be considered to be a true positive only if 75\% of its covered y-positions have a point-wise distance less than the max length (1.5 meters). The percentages of matched ground-truth lanes and matched prediction lanes are reported as recall and precision. The Average Precision (AP) and maximum F-score are reported as a comprehensive evaluation and an evaluation of the best operation point. 
For the real-world FLCP Dataset, we use the CULane's F1~\cite{SCNN} for evaluation. In addition, we also report the FPS, MACs~\cite{MACs} (one multiply-accumulate operation is approximated by two floating operations (FLOPs)), and the total number of network parameters.

\noindent \textbf{Implementation Details.}
For a fair comparison with other methods, we also set the input resolutions of the image and the top-view image to $360\times480$ and $208\times108$, respectively. The flat ground plane has the range of $X\in\left[-10, 10\right]\times Y\in\left[1, 101\right]$ meters. The same Y-position sequence $\left\{3, 5, 10, 15, 20, 30, 40, 50, 65, 80, 100\right\}$ is used to re-sample lane points since the visual appearance in the distance gets sparser in top-view. The fixed number of predicted curves $M$ and polynomial order $R$ are set as 7 and 3. 
The batch size, training iterations and learning rate are 16, 450k, and 0.0001.
$\alpha_1$, $\alpha_2$ and $\alpha_3$ are set as 1, 5, 5. 

\noindent \textbf{Baselines.}
We treat the previous state-of-the-art \textbf{Gen-LaneNet}~\cite{GenLaneNet} and \textbf{3D-LaneNet}~\cite{3DLaneNet} as competitors. Their results are dependent on perfect camera poses provided by the dataset. \textbf{CLGo} (\textbf{C}amera pose and 3D \textbf{L}ane with \textbf{G}ometry c\textbf{o}nstraints) is the final proposed method integrating Stage 1 and 2, which gets rid of ground truth camera poses during evaluation. Meanwhile, the \textbf{Stage 1} is also engaged because of no need perfect poses either.

\subsection{Comparisons with State-of-the-Art Methods}
In tab.~\ref{tab:OverallResult}, our CLGo achieves the highest F-Score and AP for all scenes without ground truth camera poses, and outperforms previous SOTA Gen-LaneNet for a large margin---\textbf{3.8}\% and \textbf{4.1}\% in balanced scene, \textbf{8.1}\% and \textbf{9.3}\% in rarely observed scene, \textbf{2.0}\% and \textbf{2.0}\% in scene with visual variations. Comparison results prove the effectiveness of the proposed methods for jointly learning camera pose and 3D lanes from a monocular camera. 
We notice that our local accuracy is comparable or a bit worse than SOTA, especially along Z-axis. This is because the fitting procedure requires unified coefficients to fit all local points simultaneously. Such a strategy may not be flexible enough to localize every local point accurately but is more reliable to capture the whole lane shape than the previous method (validated by F-Score and AP performance). 
Moreover, the poorer results of Stage 1 also prove the importance of camera poses in monocular 3D lane detection to transform the image into top-view.  

\noindent \textbf{Comparison of efficiency.}
Tab.~\ref{tab:ResourceResult} demonstrates the comparison about the resource consumption. The CLGo works without ground truth camera poses and additional post-processing. Compared with Gen-LaneNet, our method consumes \textbf{2.26}$\times$ fewer parameters, \textbf{19.8}$\times$ less computation complexity and runs \textbf{1.25} $\times$ faster than it.

\noindent \textbf{Qualitative comparisons.}
The visualization of the lane detection results is given in Fig.~\ref{fig:qual3dlane}. The left three columns demonstrate our method performs accurate fitting results, especially at the distant areas. As for the right three columns, our methods show complete and accurate lane estimations while the prediction of anchor-based methods was either incomplete (middle), or missed a lane entirely and wrongly clustered which causes lane crossing (top). Reliably perceiving lane structures completely is essential to keep driving safe, such as avoiding unwanted lane changes. We attribute improvements to (1) representing lanes as polynomials preserve smooth lane structures and embed global continuity to help fit the whole lane; (2) extracting non-local features is vital to learn lanes' long and thin structures for improving fitting performance, especially at remote regions.

\begin{figure}[t]
\begin{center}
\includegraphics[width=83mm]{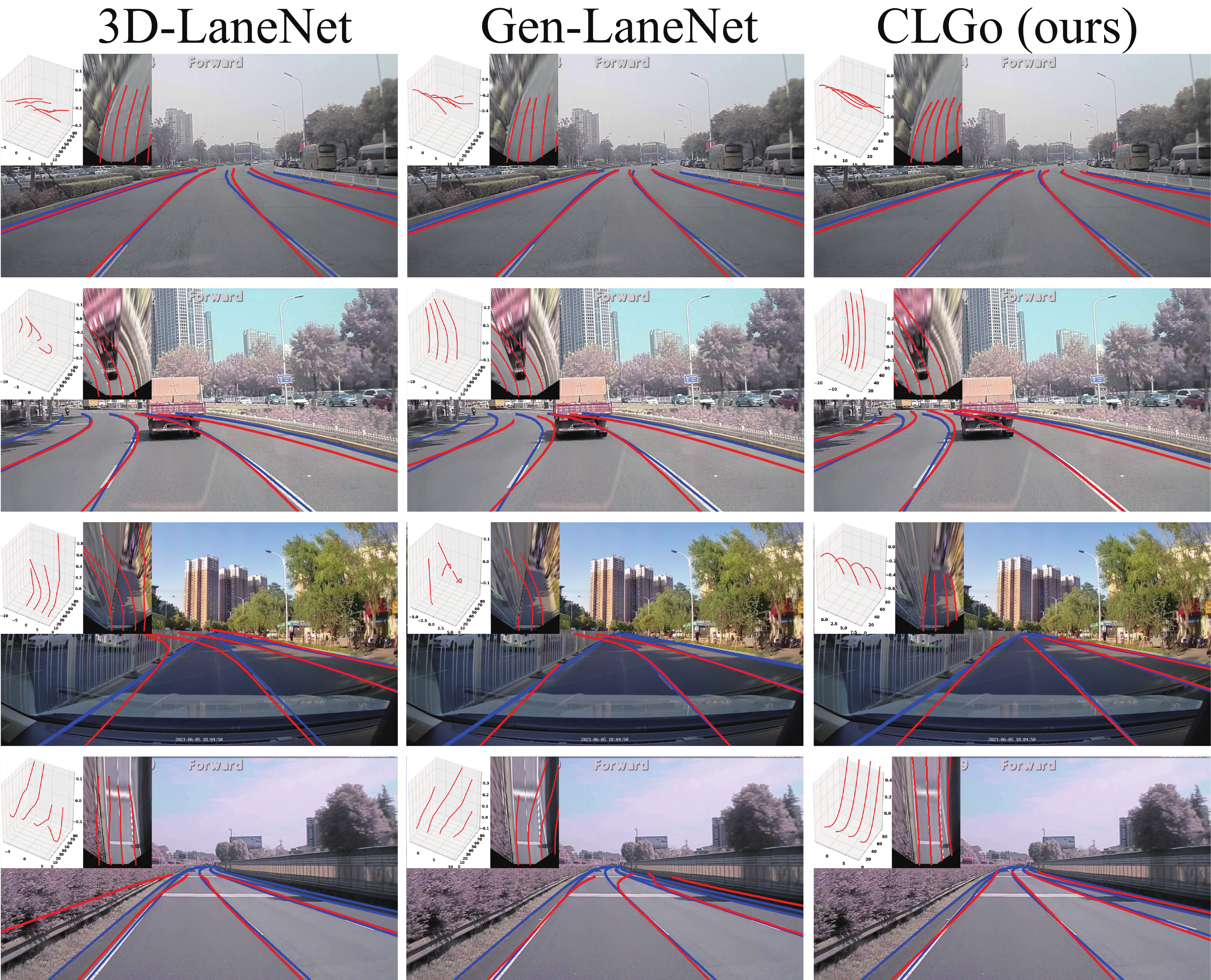}
\end{center}
\caption{\textbf{Qualitative comparisons} on the self-collected real-world FLCP dataset. Red and blue lines indicate the estimation and ground truth lanes respectively.}
\label{fig:flcp-demo}
\end{figure}
\begin{table}[t]
\begin{center}
\setlength{\tabcolsep}{3.4mm}{
\begin{tabular}{|l|c|c|c|c|}
\hline
Method & CP      & F1 & Prec. & Rec.  \\
\hline
\hline
3D-LaneNet   & GT      & 19.28   & 24.45 & 15.92 \\
Gen-LaneNet  & GT      & 30.87   & 40.72 & 24.86 \\
Stage 1~(ours)      & $\backslash$ & 21.53   & 25.71 & 18.52 \\
CLGo~(ours)         & PD      & \textbf{33.82}   & \textbf{40.80} & \textbf{28.88} \\
\hline
\end{tabular}}
\caption{Comparisons on the real-world FLCP dataset~(\%).}
\label{tab:real-studies}
\end{center}
\vspace{-0.5em}
\end{table}
\noindent \textbf{Evaluation on real-world images.}
Results in Tab.~\ref{tab:real-studies} show that our CLGo also has superior performance. The Stage 1 results are still poor, which validates the importance of using camera parameters to transform the viewpoint for realistic data. 
Fig.~\ref{fig:flcp-demo} shows qualitative comparisons. Facing unseen realistic scenes, our method predicates more reasonable, smooth, and continuous 3D lanes than others. Using the same camera poses, the projected 2D lanes of CLGo are also more complete and accurate. \textit{Much more qualitative results on real scenes can be found in the appendix.}

\subsection{Ablation Study}
\begin{table}[t]
\begin{center}
\setlength{\tabcolsep}{3.6mm}{
\begin{tabular}{|l|c|c|c|c|}
\hline
Training                & CP             & F-Score & AP   & Error   \\
\hline
\hline
\multirow{3}*{\makecell[c]{Jointly}}     & GT & 87.5    & 89.4 & 0.85    \\
~                                        & PD & 87.3    & 89.2 & 0.90    \\
\cline{2-5}
~  & dc & \textbf{-0.2}    & \textbf{-0.2} & \textbf{+0.05}    \\
\hline                  
\multirow{3}*{\makecell[c]{Separately}}  & GT & 86.1    & 87.9 & 0.83    \\
~                                        & PD & 84.9    & 86.6 & 0.96    \\
\cline{2-5}
~  & dc & -1.2    & -1.3 & +0.13    \\
\hline
\end{tabular}}
\caption{Comparison between using perfect and estimated camera poses on SVV~(\%). The lower the decreased value (dc), the better the performance.}
\label{tab:viewstudy}
\end{center}
\vspace{-0.5em}
\end{table}
\noindent \textbf{Comparison between using perfect and estimated camera pose.}
To evaluate the performance of camera pose regression, we test the CLGo by using ground truth camera pose and analyze the changes.   
As Tab.~\ref{tab:viewstudy} shows, when we jointly train two stages, the CLGo shows a very severe decline, which ensures the accuracy of camera pose regression.
However, when we train stages separately, the performance shows a quite obvious degradation.
Therefore, the multi-task joint training strategy benefits the performance a lot, which mainly comes from the improved tasks' compatibility. \textit{More comparisons of previous methods among all scenes can be found in the appendix, which also shows that our jointly trained CLGo has the fewest decrease.}

\begin{table}[t]
\begin{center}
\setlength{\tabcolsep}{1.2mm}{
\begin{tabular}{|l|cccc|ccc|}
\hline
~ & \multicolumn{4}{c|}{Stage 1 Loss} & \multicolumn{3}{c|}{Stage 2 Loss}  \\
Config & $L_{cam}$ & $L_{3D}$ & $L_{top}$ & $L_{img}$ & $L_{3D}$ & $L_{top}$ & $L_{img}$ \\
\hline
\hline 
T1 & \checkmark & ~ & ~ & ~ & \checkmark & ~ & ~ \\
T2 & \checkmark & \checkmark & ~ & ~ & \checkmark & ~ & ~ \\
T3 & \checkmark & ~ & \checkmark & \checkmark & \checkmark & ~ & ~ \\
T4 & \checkmark & \checkmark & \checkmark & \checkmark & \checkmark & ~ & ~ \\
T5 & \checkmark & \checkmark & \checkmark & \checkmark & \checkmark & \checkmark & ~ \\
T6 & \checkmark & \checkmark & \checkmark & \checkmark & \checkmark & \checkmark & \checkmark \\
\hline
\hline
Result   & \multicolumn{2}{c|}{Pitch}  & \multicolumn{2}{c|}{Height}        
& F-Score & AP & Error \\
\hline
\hline
T1 & \multicolumn{2}{c|}{$\mbox{0.236}^\circ$} & \multicolumn{2}{c|}{0.041} 
& 84.5 & 86.3  & 1.10 \\
T2 & \multicolumn{2}{c|}{$\mbox{0.223}^\circ$} & \multicolumn{2}{c|}{0.033} 
& 85.5 & 87.3  & 1.00 \\
T3 & \multicolumn{2}{c|}{$\mbox{0.164}^\circ$} & \multicolumn{2}{c|}{0.027} 
& 86.3 & 88.1  & 0.96 \\
T4 & \multicolumn{2}{c|}{$\mbox{\textbf{0.155}}^\circ$} & \multicolumn{2}{c|}{0.026} 
& \textbf{87.3} & \textbf{89.2}  & \textbf{0.90} \\
T5 & \multicolumn{2}{c|}{$\mbox{0.156}^\circ$} & \multicolumn{2}{c|}{\textbf{0.025}} 
& 87.1 & 89.0  & 0.91 \\
T6 & \multicolumn{2}{c|}{$\mbox{0.159}^\circ$} & \multicolumn{2}{c|}{0.028} 
& 85.3 & 87.0  & 1.03 \\
\hline
\end{tabular}}
\caption{Evaluation of losses for two stages on SVV~(\%).}
\label{tab:ablationTab}
\end{center}
\vspace{-0.5em}
\end{table}
\noindent \textbf{Effect of geometry constraints.}
In this section, we gradually add geometry constraints to examine their contributions on two stages of CLGo. We attend to the pitch and ultimate 3D lane performances of CLGo, since the height error is extremely small in the order of magnitude ($\mbox{10}^{-2}$ centimeters), which plays a small role.
Tab.~\ref{tab:ablationTab} shows the results of the proposed variations. The F-Score and AP drop 1.8\% and 1.9\% without geometry constraints comparing T2 and T4 (1.8\% and 1.8\% comparing T1 and T3), 1.0\% and 1.1\% without the auxiliary 3D lane branch comparing T3 and T4 (1.0\% and 1.0\% comparing T1 and T2). The reason for the above degradation is the deteriorating camera pose estimation, indicating that the proposed auxiliary branch and re-projection consistencies influence the most to the whole network.

As for their effects on stage 2, the performance holds when adding constraint in the flat ground plane (T4 and T5), but with constrained points at the image plane (T6), the performance degrades significantly. We attribute the former to the top-view image already contains shape patterns about varied undulations, so such a constraint does not bring a significant effect. Meanwhile, we think the drop in the T6 is caused by viewpoint misalignment between the image plane constraint and top-view input. Many contextual clues such as skyline, trees, or buildings have been lost in the top-view, which decreases the network's ability to reconstruct lanes' constraints in the perspective-view image plane.

\section{Conclusion}
In this work, we present a full-vision-based 3D lane and camera pose estimation framework. The geometry constraints improve consistencies between 3D representations and 2D input by introducing mathematical geometry priors in model learning, as well as enhances compatibility between camera pose and 3D lane tasks, leading to significant improvements on both of them. The whole framework is validated with reliably estimated camera poses and outperforms state-of-the-art methods which are all evaluated using ground truth camera poses, while achieving the lightest model size, fewest computation costs, and the fastest FPS. 
It would be interesting to combine flexible representations for fitting lanes with complex topologies in future work.

\section{Acknowledgement}
This work was supported by the National Natural Science Foundation of China (61976170, 91648121, 62088102).

\bibliography{aaai22.bib}

\section{Appendix}
In this supplemental document, we provide:

\S \textbf{A}~Investigation of performance variation when using perfect and estimated camera pose.

\S \textbf{B}~~Comparison with monocular camera pose regression method.

\S \textbf{C}~~Investigation of constructing geometry constraints using different choices of 3D lanes.

\S \textbf{D}~~Investigation of two-stage framework variations.

\S \textbf{E}~~Effect of using Transformer.

\S \textbf{F}~~Qualitative results of different road conditions on 3D synthetic dataset.

\S \textbf{G}~~Qualitative comparisons on FLCP.

\noindent \textbf{A. Investigation of performance variation when using perfect and estimated camera pose.}

This part investigates how robust the proposed camera pose regression method can be to get camera pose for 3D lane detection.
We list four methods, 3D-LaneNet~\cite{3DLaneNet}, Gen-LaneNet~\cite{GenLaneNet}, our CLGo, and a CLGo whose two stages are trained separately (denoted as CLGo (sep.)), and compare the difference in the performance of each method when the camera pose is true and predicted.
Tab.~\ref{tab:cpsensisitivity} shows the results.
The detection performance of all detectors degrades more or less after using the estimated camera poses. However, our CLGo degrades the fewest F-Score, AP, and fitting errors, which validates the effectiveness and robustness of our camera pose regression method and multi-task jointly learning strategy. As a result, our CLGo is more applicable and robust to perceive 3D lanes when facing dynamic driving scenes.

\noindent \textbf{B. Comparison with monocular single-frame camera pose regression method.}

In this paragraph, we evaluate the camera pose errors on SVV, comparing with SOTA monocular single-frame visual camera pose methods, including PoseNet~\cite{Kendall2015ICCV}, PoseGeo~\cite{Kendall2017CVPR}, PoseLSTM~\cite{Walch2017ICCV}, since the 3D lane dataset is not sequential. We modify their public codes to adapt to our task settings.
The evaluation results shown in Table.~\ref{tab:pitch-study} indicate that our method can achieve an accurate pose as other specifically designed methods. 

\begin{table}[t]
\begin{center}
\setlength{\tabcolsep}{4.3mm}{
\begin{tabular}{|l|c|c|}
\hline
            & Height & Pitch \\
\hline
\hline
PoseNet     & 0.054  & 0.234    \\
PoseGeo     & 0.024  & 0.138    \\
PoseLSTM    & 0.034  & 0.211    \\
CLGo        & 0.026  & 0.155    \\
\hline
\end{tabular}}
\caption{Quantitative comparisons of monocular single-frame camera pose regression method. Our CLGo achieves an accurate result as other specifically designed pose estimators.}
\label{tab:pitch-study}
\end{center}
\vspace{-1.5em}
\end{table}

\begin{figure*}[t]
\begin{center}
\includegraphics[width=177mm]{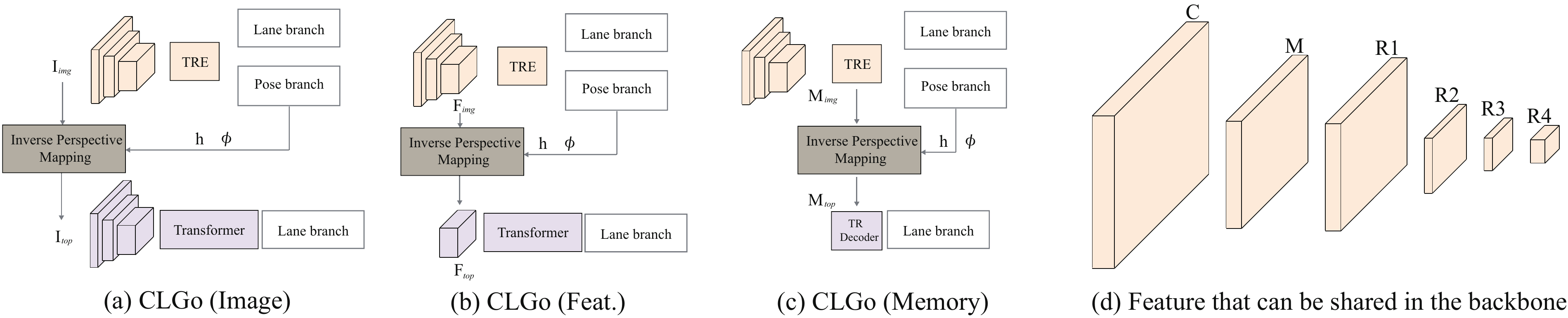}
\end{center}
\caption{Demonstration of structure variations. The IPM is a differential operation, so applying it on features instead of the image could make the parameters of network before the IPM hook to be reused for two stages.}
\label{fig:investigation-structure}
\end{figure*}

\begin{figure}[h]
\begin{center}
\includegraphics[width=83mm]{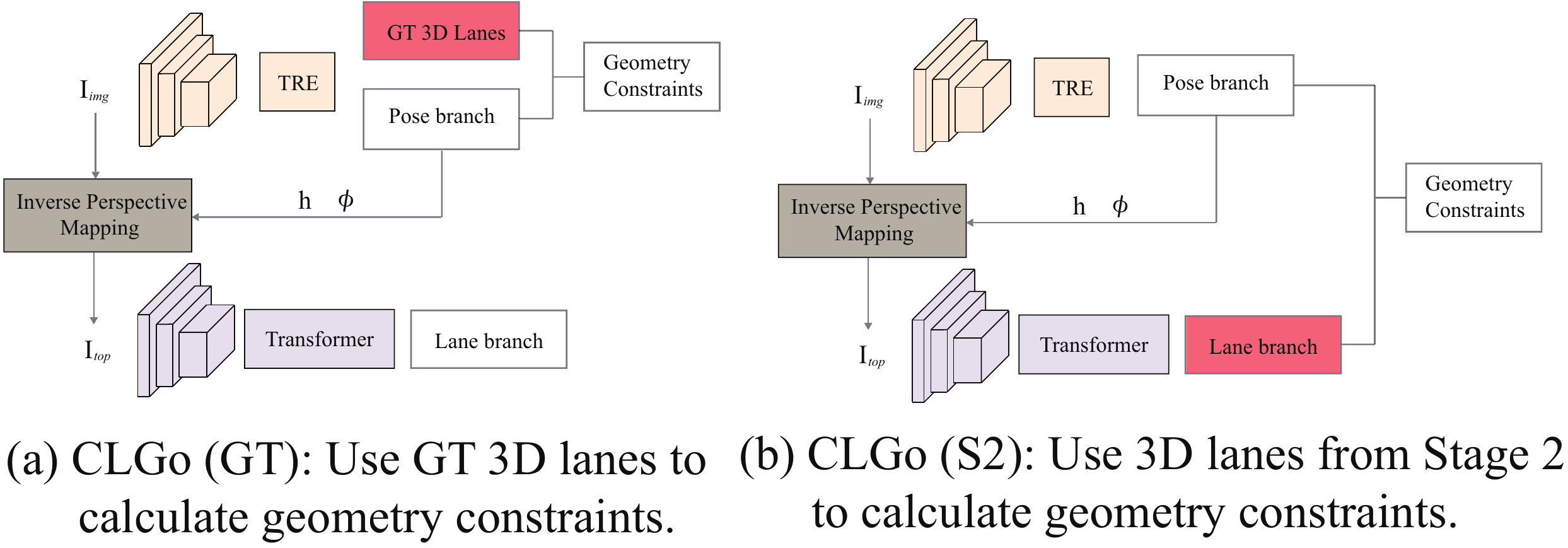}
\end{center}
\caption{Constructing geometry constraints with different choices of 3D lanes. Different from the proposed CLGo network, (a) uses the ground truth 3D lanes. (b) uses the 3D lanes estimated from Stage 2.}
\label{fig:investigation-lanes}
\end{figure}

\begin{table}[h]
\begin{center}
\setlength{\tabcolsep}{1.8mm}{
\begin{tabular}{|l|c|c|c|c|c|}
\hline
          & Height & Pitch & F-Score & AP   & Error   \\
\hline
\hline
CLGo      & 0.026  & 0.155 & 87.3    & 89.2 & 0.90    \\
CLGo~(GT) & 0.027  & 0.164 & 86.3    & 88.1 & 0.96    \\
CLGo~(S2) & 0.030  & 0.198 & 85.4    & 87.1 & 1.05    \\
\hline
\end{tabular}}
\caption{Quantitative results of different 3D lanes for constructing geometry constraints.}
\label{tab:lane-study}
\end{center}
\vspace{-1.5em}
\end{table}

\noindent \textbf{C. Investigation of constructing geometry constraints using different choices of 3D lanes.}

This section investigates which 3D lane choice is the best for constructing geometry constraints. To do so, we compare CLGo with two other options. The structure of Fig.~\ref{fig:investigation-lanes}~(a) uses the ground truth 3D lanes to calculate geometry constraints, denoted as CLGo~(GT). The other structure of Fig.~\ref{fig:investigation-lanes}~(b) uses the estimation of 3D lanes from Stage 2 to calculate geometry constraints, denoted as CLGo~(S2). The proposed CLGo uses the estimation of 3D lanes from Stage 1.
In Tab.~\ref{tab:lane-study}, CLGo~(GT) and CLGo~(S2) both have lower performance than CLGo. 
We attribute it to (1): CLGo~(GT) brings geometry consistencies from ground truth lanes is not flexible enough for model learning, leading to a high risk for overfitting; (2): CLGo~(S2) calculates geometry constraints using lanes from Stage 2 and the camera pose from Stage 1, making gradients of geometry constraints back-propagated through the top-view pipeline. However, the top-view image mainly contains lanes appearance and already drops many contextual clues such as skyline, trees and traffic objects. In contrast, those clues are included in the perspective-view image and crucial feature for a 3D task, \emph{e.g.,} the skyline for 3D camera task~\cite{MonoEF}. Therefore, gradients of geometry constraints back-propagated through the top-view pipeline will cause mis-alignment between objectives and appearances and increase reconstruction burdens, resulting in poorer performance.

\begin{table}[t]
\begin{center}
\setlength{\tabcolsep}{2.4mm}{
\begin{tabular}{|l|c|c|c|c|c|}
\hline
CLGo       & F-Score & AP   & Error   & \#Para & MACs \\
\hline
\hline
Image      & \textbf{87.3}    & \textbf{89.2} & \textbf{0.90}    & 1.528  & 0.497 \\
Feat.C     & 87.2    & 89.1 & 0.93    & 1.525  & 0.481 \\
Feat.M     & 86.7    & 88.7 & 0.95    & 1.525  & 0.481 \\
Feat.R1    & 87.0    & 88.9 & 0.92    & 1.521  & 0.474 \\
Feat.R2    & 86.7    & 88.7 & 1.01    & 1.488  & 0.460 \\
Feat.R3    & 87.2    & 89.1 & 1.03    & 1.356  & 0.446 \\
Feat.R4    & 86.8    & 88.8 & 1.15    & 0.830  & 0.432 \\
Memory     & 86.5    & 88.8 & 1.13    & \textbf{0.805}  & \textbf{0.431} \\
\hline
\end{tabular}}
\caption{Quantitative results of different two-stage architectures on SVV~(\%). The Error means total lane fitting error given in meters.}
\label{tab:stage-study}
\end{center}
\vspace{-0.05em}
\end{table}

\noindent \textbf{D. Investigation of two-stage structure variations.}

This study investigates whether the two-stage structure can share features and how sharing features perform.  
Since the \textit{Inverse projective mapping} (IPM) is a differential operator that can transform images and features, we would apply it on features, making the network parameters before the sharing position would be reused for two stages.
Fig.~\ref{fig:investigation-structure} shows the variations.
Fig.~\ref{fig:investigation-structure}~(a) is the proposed no sharing structure, denoted as CLGo~(Image), where the Image means applying IPM on images.
Fig.~\ref{fig:investigation-structure}~(b) is feature sharing structure, and Fig.~\ref{fig:investigation-structure}~(d) shows six choices of features to use IPM. For example, applying IPM on the output features of backbone to make the whole backbone shared for two stages will be denoted as CLGo~(Feat.R4).
Fig.~\ref{fig:investigation-structure}~(c) is a memory sharing structure, denoted as CLGo~(Memory), where the Memory means applying IPM on Transformer encoder's output. In Tab.~\ref{tab:stage-study}, sharing parameters can reduce the number of parameters and computation costs, but it inevitably produces degradation on detection performance, especially for the localization errors.
This tells us that features learned from two viewpoints can not be shared well. However, applying sharing structure is still applicable to mobile platform which have limited resources, since the performance only degrades a few (also validates the scalability of our CLGo).

\noindent \textbf{E. Effect of Transformer.}

\begin{figure}[t]
\begin{center}
\includegraphics[width=83mm]{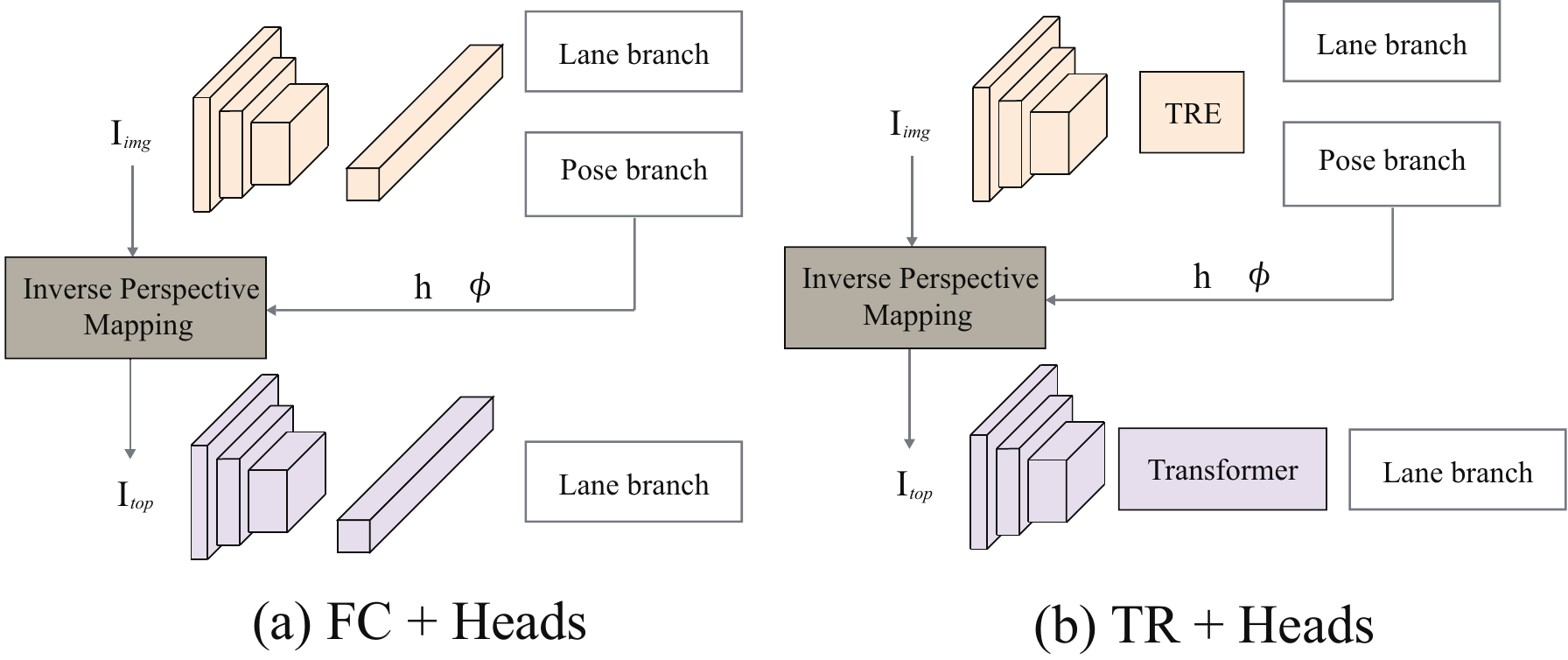}
\end{center}
\caption{{Demonstration of head variations}.}
\label{fig:investigation-head}
\end{figure}

In this part, we study the effect of the Transformer. We replace the transformer with a similar-size fully-connected layers as Fig.~\ref{fig:investigation-head} illustrates. Tab.~\ref{tab:head-study} shows that without using transformer, the height and pitch error raises 0.062 and 0.183, the global accuracy decreases by 6.2\% and 5.5\% in terms of F-score and AP, and the localization error raises 0.47 meters. The results indicate transformer is better than fully-connected layers due to its non-local modeling ability to capture long and thin lane structures as well as global context aggregation ability about scene understanding for accurate camera pose regression.

\begin{table}[t]
\begin{center}
\setlength{\tabcolsep}{1.9mm}{
\begin{tabular}{|l|c|c|c|c|c|}
\hline
           & Height & Pitch & F-Score & AP   & Error   \\
\hline
\hline
FC + Heads & 0.088 & 0.338 & 81.1    & 83.7 & 1.37    \\
TR + Heads & 0.026 & 0.155 & 87.3    & 89.2 & 0.90    \\
\hline
\end{tabular}}
\caption{Quantitative results of transformer-based head or fully-connected-layers-based head. Without transformer, the performance decreases obviously.}
\label{tab:head-study}
\end{center}
\vspace{-0.5em}
\end{table}

\begin{table*}[h]
\begin{center}
\setlength{\tabcolsep}{3.1mm}{
\begin{tabular}{|c|l|c|c|c|c|c|c|c|}
\hline
Scene & Method & CP & F-Score & AP & X error near & X error far & Z error near & Z error far \\
\hline
\hline
\multirow{12}*{\makecell[c]{BS}} & \multirow{3}*{\makecell[c]{3D-LaneNet}} & GT
                 & 86.4 & 89.3 & 0.068 & 0.477 & 0.015 & 0.202 \\
~ & ~ & PD       & 84.9 & 87.4 & 0.072 & 0.508 & 0.035 & 0.264 \\
\cline{3-9}
~ & ~ & decrease & -1.5 & -1.9 & 0.004 & 0.031 & 0.020 & 0.062 \\
\cline{2-9}

~ & \multirow{3}*{\makecell[c]{Gen-LaneNet}} & GT 
                 & 88.1 & 90.1 & 0.061 & 0.496 & 0.012 & 0.214 \\
~ & ~ & PD       & 86.1 & 87.9 & 0.068 & 0.543 & 0.040 & 0.297 \\
\cline{3-9}
~ & ~ & decrease & -2.0 & -2.2 & 0.007 & 0.047 & 0.028 & 0.083 \\
\cline{2-9}

~ & \multirow{3}*{\makecell[c]{CLGo~(sep.)}} & GT 
                 & 90.7 & 92.6 & 0.081 & 0.414 & 0.019 & 0.235 \\
~ & ~ & PD       & 89.5 & 91.3 & 0.091 & 0.450 & 0.041 & 0.281 \\
\cline{3-9}
~ & ~ & decrease & -1.2 & -1.3 & 0.010 & 0.036 & 0.022 & 0.046 \\
\cline{2-9}

~ & \multirow{3}*{\makecell[c]{CLGo}} & GT 
                 & 91.9 & 94.1 & 0.070 & 0.388  & 0.027  & 0.245 \\
~ & ~ & PD       & 91.7 & 93.9 & 0.071 & 0.386  & 0.036  & 0.253 \\
\cline{3-9}
~ & ~ & decrease & \textbf{-0.2} & \textbf{-0.2} & \textbf{0.001} & \textbf{-0.002} & \textbf{-0.009} & \textbf{-0.008} \\

\hline
\hline
\multirow{12}*{\makecell[c]{ROS}} & \multirow{3}*{\makecell[c]{3D-LaneNet}} & GT
                 & 72.0 & 74.6 & 0.166 & 0.855 & 0.039 & 0.521 \\
~ & ~ & PD       & 65.8 & 67.6 & 0.176 & 0.944 & 0.080 & 0.641 \\
\cline{3-9}
~ & ~ & decrease & -6.2 & -7.0 & 0.010 & 0.089 & 0.041 & 0.120 \\
\cline{2-9}

~ & \multirow{3}*{\makecell[c]{Gen-LaneNet}} & GT 
                 & 78.0 & 79.0 & 0.139 & 0.903 & 0.030 & 0.539 \\
~ & ~ & PD       & 72.5 & 73.3 & 0.156 & 1.010 & 0.087 & 0.711 \\
\cline{3-9}
~ & ~ & decrease & -5.5 & -5.7 & 0.017 & 0.107 & 0.057 & 0.172 \\
\cline{2-9}

~ & \multirow{3}*{\makecell[c]{CLGo~(sep.)}} & GT 
                 & 85.7 & 87.8 & 0.166 & 0.793 & 0.046 & 0.578 \\
~ & ~ & PD       & 79.7 & 81.4 & 0.207 & 0.860 & 0.092 & 0.661 \\
\cline{3-9}
~ & ~ & decrease & -6.0 & -6.4 & 0.041 & 0.067 & 0.046 & 0.097 \\
\cline{2-9}

~ & \multirow{3}*{\makecell[c]{CLGo}} & GT 
                 & 88.3 & 90.7 & 0.133 & 0.733  & 0.041 & 0.589 \\
~ & ~ & PD       & 86.1 & 88.3 & 0.147 & 0.735  & 0.071 & 0.609 \\
\cline{3-9}
~ & ~ & decrease & \textbf{-2.2} & \textbf{-2.4} & \textbf{0.014} & \textbf{0.002} & \textbf{0.030} & \textbf{0.020} \\

\hline
\hline
\multirow{12}*{\makecell[c]{SVV}} & \multirow{3}*{\makecell[c]{3D-LaneNet}} & GT 
                 & 72.5 & 74.9 & 0.115 & 0.601 & 0.032 & 0.230 \\
~ & ~ & PD       & 71.4 & 73.1 & 0.121 & 0.618 & 0.055 & 0.294 \\
\cline{3-9}
~ & ~ & decrease & -1.1 & -1.8 & 0.006 & 0.017 & 0.023 & 0.064 \\
\cline{2-9}

~ & \multirow{3}*{\makecell[c]{Gen-LaneNet}} & GT 
                 & 85.3 & 87.2 & 0.074 & 0.538 & 0.015 & 0.232 \\
~ & ~ & PD       & 82.3 & 84.0 & 0.091 & 0.614 & 0.058 & 0.349 \\
\cline{3-9}
~ & ~ & decrease & -3.0 & -3.2 & 0.017 & 0.076 & 0.043 & 0.117 \\
\cline{2-9}

~ & \multirow{3}*{\makecell[c]{CLGo~(sep.)}} & GT 
                 & 86.1 & 87.9 & 0.096 & 0.449 & 0.027 & 0.257 \\
~ & ~ & PD       & 84.9 & 86.6 & 0.103 & 0.501 & 0.050 & 0.308 \\
\cline{3-9}
~ & ~ & decrease & -1.2 & -1.3 & 0.007 & 0.052 & 0.023 & 0.051 \\
\cline{2-9}

~ & \multirow{3}*{\makecell[c]{CLGo}} & GT 
                 & 87.5 & 89.4 & 0.083  & 0.456  & 0.030 & 0.279 \\
~ & ~ & PD       & 87.3 & 89.2 & 0.084  & 0.464  & 0.045 & 0.312 \\
\cline{3-9}
~ & ~ & decrease & \textbf{-0.2} & \textbf{-0.2} & \textbf{0.001} & \textbf{0.008} & \textbf{0.015} & \textbf{0.033} \\
\hline
\end{tabular}}
\caption{Comparisons about camera pose sensitivity on 3D lane synthetic dataset testing set~(\%). The X error and Z error are given in \textbf{meters}. CP, GT and PD are abbreviations of camera pose, ground truth and prediction. The decrease calculate the performance drops after using the same prediction camera poses. The lower the decreased value, the better the performance.}
\label{tab:cpsensisitivity}
\end{center}
\end{table*}

\noindent \textbf{F. Qualitative results of uphill, downhill, and flat roads.}

Additional qualitative results on the public 3D Synthetic Lane dataset are shown in Fig.~\ref{fig:qual3dlane}. We show the appearance patterns on different road undulations to visualize the insights of our geometry constraints.
\begin{figure*}[t]
\begin{center}
\includegraphics[width=174mm]{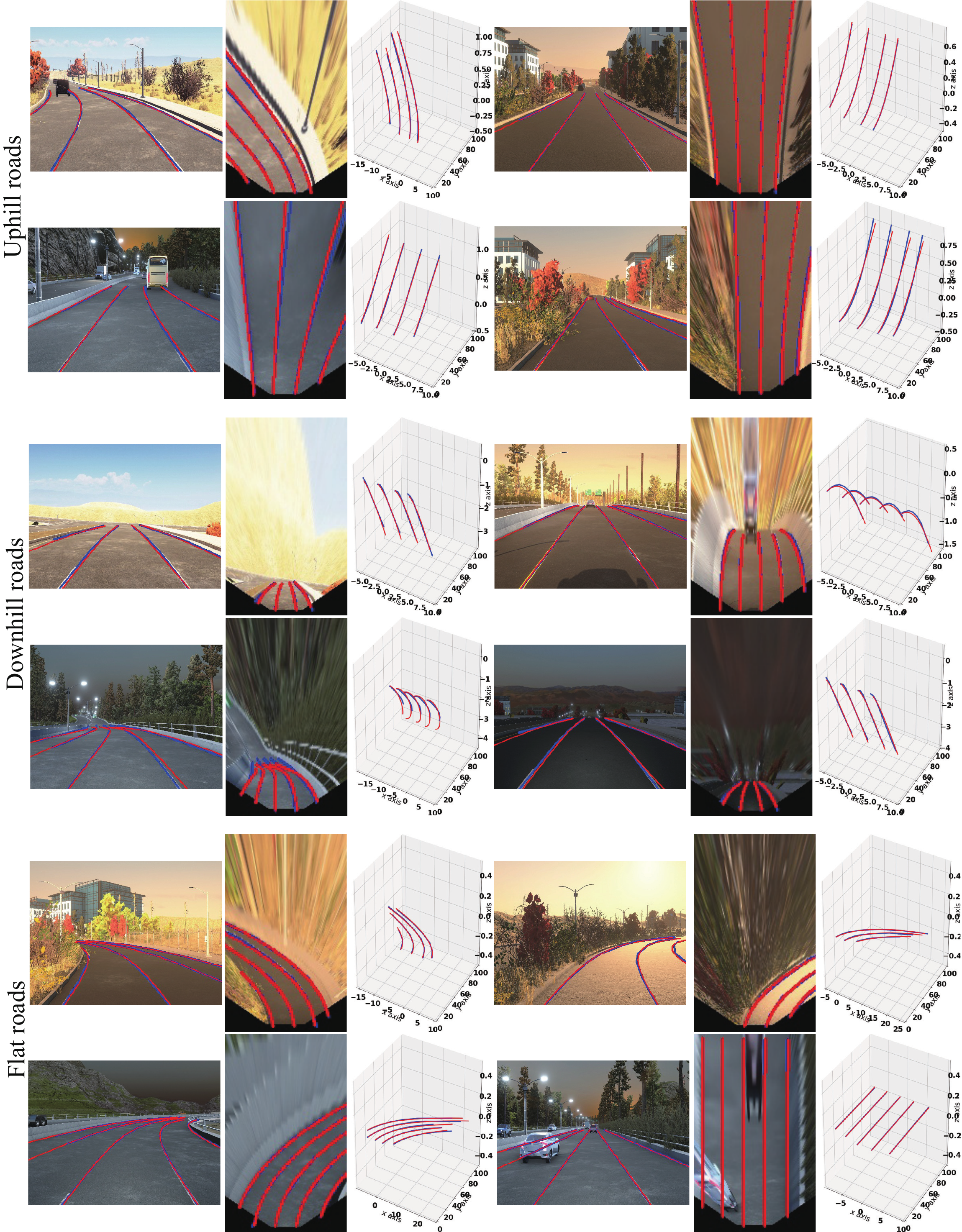}
\end{center}
\caption{{Qualitative results on different road conditions}. Red and blue lines indicate the estimation and ground truth lanes respectively. From right to left, we sequentially show the 3D fitting results and their projections on the flat ground plane and image plane.}
\label{fig:qual3dlane}
\end{figure*}

\noindent \textbf{G. Qualitative comparisons on FLCP dataset.}

\begin{figure*}[t]
\begin{center}
\includegraphics[width=174mm]{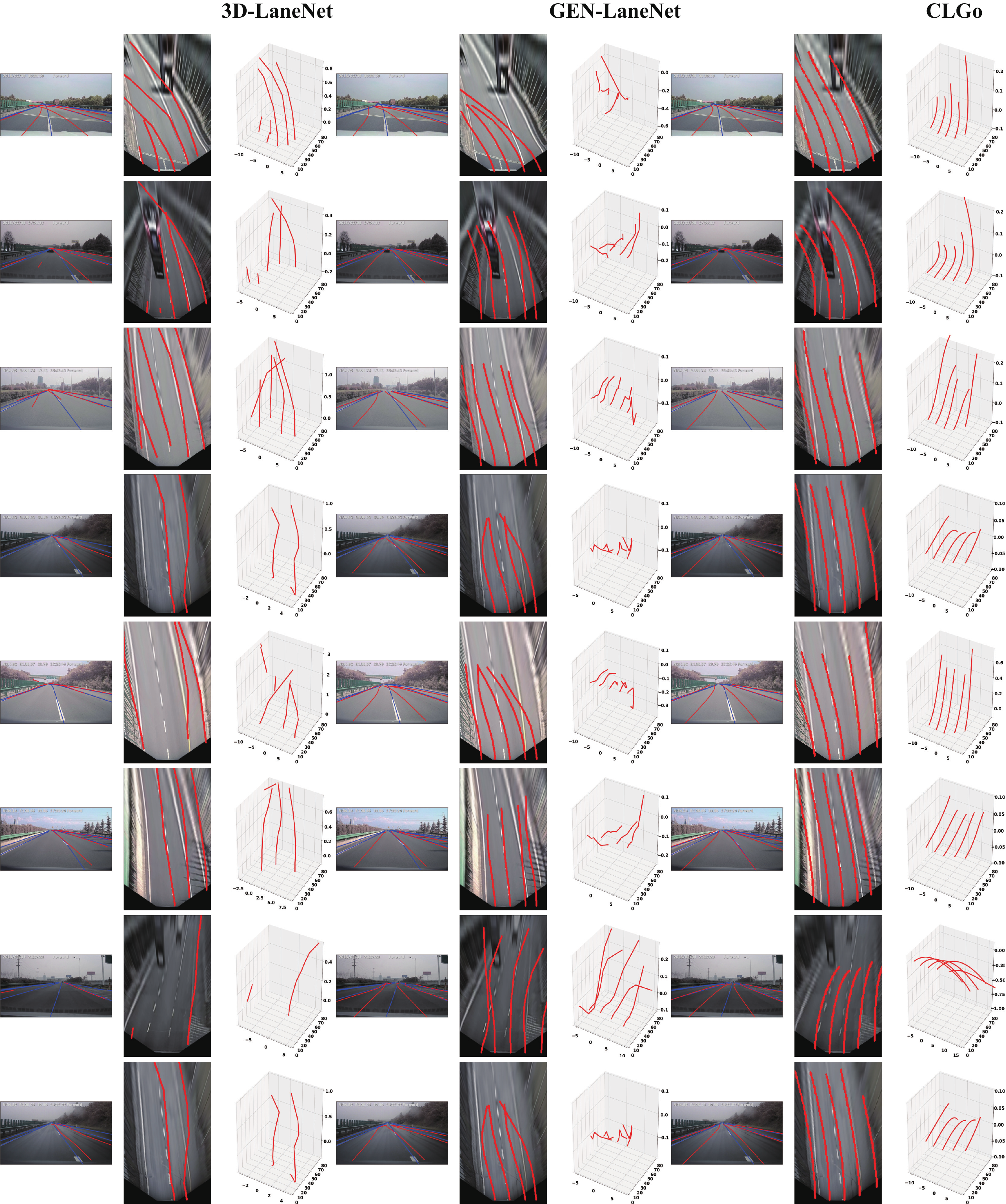}
\end{center}
\caption{{Qualitative comparisons on FLCP dataset}. From right to left, we sequentially show the 3D fitting results and their projections on the flat ground plane and image plane. Our CLGo is more adaptable to realistic images by more complete, smooth, and continuous 3D lane detections as well as fewer false crossing, missing, and broken 3D lane detections}
\label{fig:qual3dlane-flcp}
\end{figure*}

\begin{figure*}[t]
\begin{center}
\includegraphics[width=174mm]{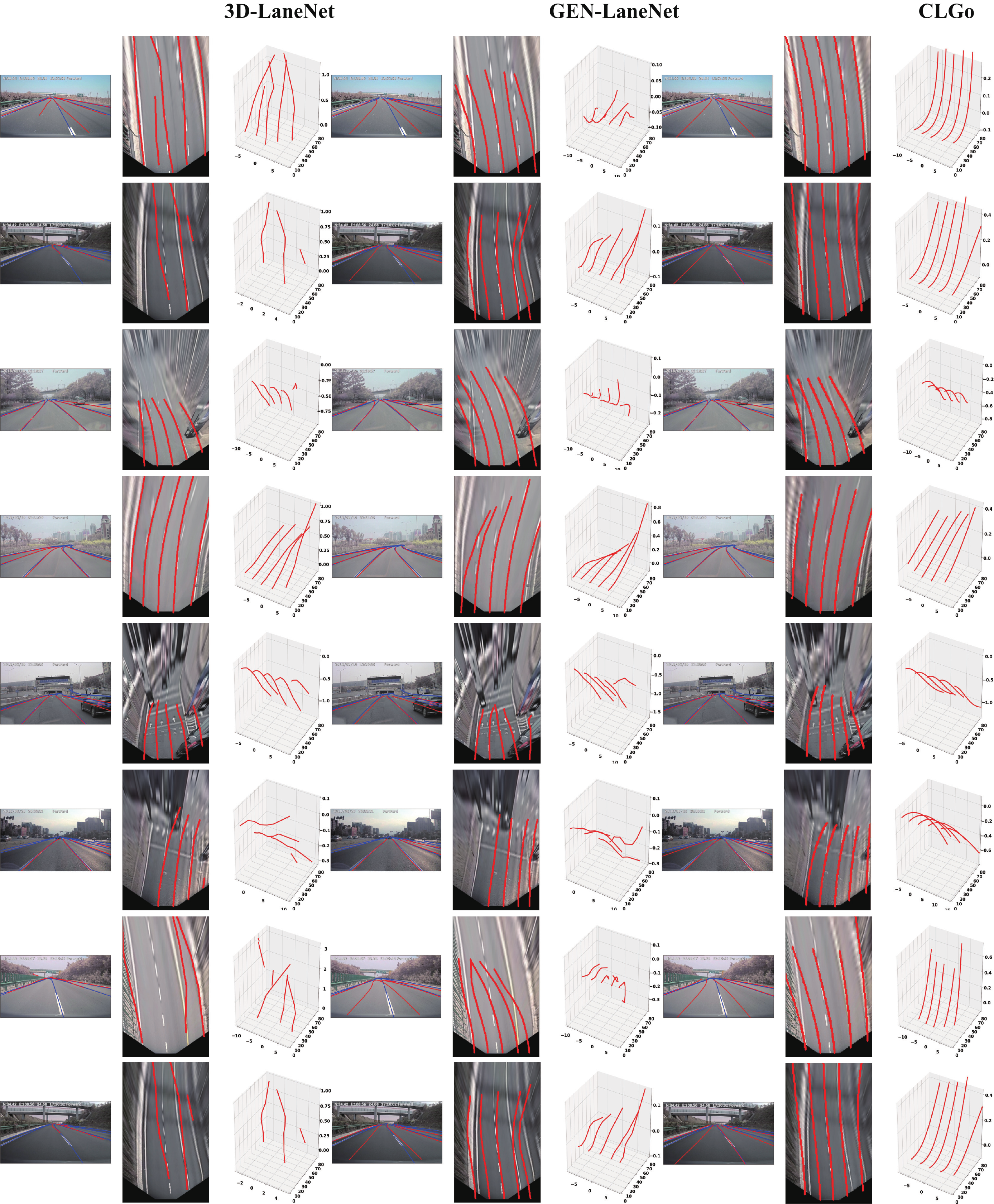}
\end{center}
\caption{{Qualitative comparisons on FLCP dataset}. From right to left, we sequentially show the 3D fitting results and their projections on the flat ground plane and image plane.Our CLGo is more adaptable to realistic images by more complete, smooth, and continuous 3D lane detections as well as fewer false crossing, missing, and broken 3D lane detections}
\label{fig:qual3dlane-flcp1}
\end{figure*}

Additional qualitative comparisons on the self-collected FLCP dataset are shown in Fig.~\ref{fig:qual3dlane-flcp} and Fig.~\ref{fig:qual3dlane-flcp1}. The test pipeline is to apply 3D lane detectors on FLCP images. The 3D lane results would be re-projected on to image plane based on camera poses. The superior performance of our method on real-world images is illustrated in three aspects. (1) For each lane, our prediction is more complete. (2) For all lanes on the road, our method has fewer false negatives. (3) As for different road conditions, our method estimates reasonable, smooth, holistic, and independent lanes. However, Gen-LaneNet~\cite{GenLaneNet} and 3D-LaneNet~\cite{3DLaneNet} predict 3D lanes outrageously. The biggest issue with their approach is that they are unstable. Their detected points or segments are too local to form a holistic lane due to a lack of heuristic global context, leading to false crossings, wrong topologies, and broken lanes.

\end{document}